\documentclass[letterpaper,journal]{IEEEtran}
\usepackage{amsmath,amsfonts}
\usepackage{algorithmic}
\usepackage{algorithm}
\usepackage{array}
\usepackage[caption=false,font=normalsize,labelfont=sf,textfont=sf]{subfig}
\usepackage{textcomp}
\usepackage{stfloats}
\usepackage{url}
\usepackage{verbatim}
\usepackage{graphicx}
\usepackage{cite}
\usepackage{float}
\usepackage[hidelinks]{hyperref}
\usepackage{booktabs}
\usepackage{threeparttable}
\usepackage{xcolor}
\usepackage{eso-pic}
\begin{document}

\bstctlcite{IEEEexample:BSTcontrol}

\title{Large Language Model-Driven Cooperative Operator Ensemble Evolution for Permutation Flow Shop Scheduling}
\author{}
\author{Rui~Xu*,~\IEEEmembership{Member,~IEEE},
       Yufan~Liao*,~\IEEEmembership{Student Member,~IEEE},
       Haoze~Lv*,~\IEEEmembership{Student Member,~IEEE},
       \\
       Shengcai~Liu,~\IEEEmembership{Member,~IEEE},
       Yi~Mei,~\IEEEmembership{Senior Member,~IEEE},
       Ke Tang,~\IEEEmembership{Fellow,~IEEE}
       
  \thanks{
  	Rui Xu and Yufan Liao are with the School of Business, Hohai University, Nanjing 211100, China
    (email: 241320010040@hhu.edu.cn; rxu@hhu.edu.cn).
    
  	Yi Mei is with the School of Engineering and Computer Science, Victoria University of Wellington, Wellington 6140, New Zealand (email: yi.mei@vuw.ac.nz).
    
  	Haoze Lv, Shengcai Liu, and Ke Tang are with the Guangdong Provincial Key Laboratory of Brain-inspired Intelligent Computation, Department of Computer Science and Engineering, Southern University of Science and Technology, Shenzhen 518055, China. Haoze Lv is also with Zhongguancun Academy, China (email: 12332421@mail.sustech.edu.cn; liusc3@sustech.edu.cn; tangk3@sustech.edu.cn).

    Corresponding author: Shengcai Liu.
    
    * equal contribution.  
}
}

\markboth{IEEE TRANSACTIONS ON EVOLUTIONARY COMPUTATION,~Vol.~XX, No.~X, XX~XXXX}%
{Xu \MakeLowercase{\textit{et al.}}: Large Language Model-Driven Cooperative Operator Ensemble Evolution for PFSP}

\maketitle

% First-page copyright footnote for the accepted manuscript, placed in the
% bottom margin to preserve the original text layout and pagination.
\AddToShipoutPictureFG*{%
  \AtPageLowerLeft{%
    \put(\LenToUnit{\dimexpr1in+\oddsidemargin\relax},\LenToUnit{12pt}){%
      \parbox[b]{\textwidth}{\footnotesize\copyright~2026 IEEE. Personal use of this material is permitted. Permission from IEEE must be obtained for all other uses, in any current or future media, including reprinting/republishing this material for advertising or promotional purposes, creating new collective works, for resale or redistribution to servers or lists, or reuse of any copyrighted component of this work in other works.}%
    }%
  }%
}

\begin{abstract}
The permutation flow shop scheduling problem (PFSP) is a classical NP-hard optimization problem in intelligent manufacturing.
The iterated greedy (IG) metaheuristic is widely adopted for solving PFSP, due to its simplicity and strong empirical performance.
However, classical IG’s single fixed destruction operator often limits exploration and causes search stagnation on large, complex instances.
This work proposes IG-DOE, a multi-operator IG algorithm that enhances exploration by switching among heterogeneous destruction operators along a single search trajectory.
Its core mechanism, stagnation-triggered sequential switching, activates the next destruction operator in an ordered destruction operator ensemble (DOE) upon stagnation, enriching the perturbation behavior of classical IG.
Moreover, to reduce reliance on expert-crafted operators, a large language model (LLM)-assisted framework, termed stagewise cooperative operator ensemble evolution (SCOE), is introduced to automatically construct a high-quality DOE through stagewise evolution, state-awareness, and cooperative evaluation.
Experiments on the challenging \textit{VRF-hard-large} benchmark, named after Vallada, Ruiz, and Framinan, show that the DOE evolved from smaller instances generalizes well to larger instances.
Under the same CPU-time limit, IG-DOE obtained much better average performance than the Q-learning iterated greedy algorithm (QIG), a state-of-the-art IG algorithm.
Further experiments on real-world industrial-data-derived instances show that the evolved DOE generalizes effectively across data distributions.
\end{abstract}


\section{Introduction}
\IEEEPARstart{S}{hop} scheduling is a fundamental task in manufacturing and has been extensively studied in operations research and intelligent optimization~\cite{fernandez2017new}.
Among the many shop scheduling models, the permutation flow shop scheduling problem (PFSP) has received significant attention because of its widespread applications in assembly lines, electronics manufacturing, and process industries~\cite{de2025systematic}.
Furthermore, the PFSP holds high theoretical value because it serves as the foundation for many practical shop scheduling variants~\cite{wu2025balance, yu2025self}.
The objective of the PFSP is to determine the optimal processing sequence of $n$ jobs on $m$ machines to minimize the makespan~\cite{PFSPNP1:johnson1954optimal}.
Since the PFSP is NP-hard~\cite{kan1976machine}, exact algorithms are computationally prohibitive for medium-to-large PFSP instances.

Consequently, many metaheuristic algorithms~\cite{Intro_IG_V1:ruiz2007simple, kurdi2020memetic} have been developed to obtain high-quality solutions for the PFSP within a reasonable computational time.
Among them, the iterated greedy (IG) algorithm has emerged as a dominant method, due to its fast convergence and high solution quality~\cite{Intro_IG_V1:ruiz2007simple,FernandezViagas2019}.
In a standard IG iteration, the current solution is perturbed in the destruction phase, reconstructed in the construction phase, and then refined by a local search procedure.
The local search procedure performs neighborhood exploitation to improve the solution, while the destruction phase generates perturbations to help the search escape local optima.
However, the classical IG algorithm typically uses a single destruction operator.
This restricts the perturbation pattern, making the algorithm unable to adapt to different stages of the search process.
As a result, the search often becomes trapped in local optima in later stages of the optimization process~\cite{QIG:karimi2023learning}.

To mitigate this limitation, researchers often incorporate multiple search operators~\cite{PeiMLZY25_AOSSurvey,ZhengLO25,GuoMZZCD25_LANSVRP}.
Indeed, multi-operator strategies have proven effective in various shop scheduling variants~\cite{BrandimarteFadda2024_RVNS_JITJSP, Jomaa2021_VNS_PM_PFSP, Xu2024_HQPSOVNS_FJSP}.
Specifically for the PFSP, the Q-learning iterated greedy algorithm (QIG) improves IG through reinforcement-learning-based control of the destruction size $d$, i.e., the number of jobs removed from the solution during destruction~\cite{QIG:karimi2023learning}.
Upon its introduction, QIG achieved state-of-the-art performance on the challenging \mbox{\textit{VRF-hard-large}} benchmark~\cite{VRF}, named after Vallada, Ruiz, and Framinan.
However, this adaptation remains within a single destruction-operator family.
That is, it only changes how many jobs are removed, but not the removal logic itself.
From a fitness-landscape perspective, varying the destruction size alone is often insufficient to escape deep basins of attraction, because the underlying topological structure of the neighborhood remains unchanged~\cite{schiavinotto2007review}.
Instead, escaping such stagnant regions requires switching among heterogeneous operators to reshape the local search space~\cite{mladenovic1997variable}.

Following this idea, we propose \mbox{IG-DOE}, a multi-operator IG algorithm built around a \textit{destruction operator ensemble} (DOE), i.e., an ordered collection of distinct destruction operators coordinated along a single search trajectory.
IG-DOE employs a stagnation-triggered sequential switching strategy that activates different destruction operators in the DOE when the search process ceases to improve.
In this way, it enriches the perturbation behavior of classical IG without altering its efficient construction and local search backbone.

However, manually designing such a DOE requires substantial expertise, because the operators should not only be effective individually but also complement each other during the search.
Recent advances in large language models (LLMs) have opened a new avenue for automatic heuristic/operator design (LLM-AHD), which can reduce the reliance on expert-crafted operators.
Representative methods such as FunSearch~\cite{FunSearch:romera2024mathematical}, ReEvo~\cite{ReEvo:ye2024reevo}, and EoH~\cite{EOH:liu2024evolution} do not construct complete algorithms from scratch.
Instead, they evolve key heuristic components within predefined algorithmic frameworks by iteratively prompting the LLM to generate new candidates based on previous high-performing heuristics.

LLM-AHD offers a promising way to automatically construct a DOE.
However, existing methods often optimize individual operators, whose standalone performance does not necessarily translate into improved ensemble performance.
DOE construction therefore requires operator generation and evaluation to account for the current ensemble and its coordinated execution.

To address this challenge, we propose the stagewise cooperative operator ensemble evolution (SCOE) framework.
Its incremental construction strategy (ICS) builds the DOE one operator at a time.
Through state-awareness and cooperative evaluation, SCOE generates candidates informed by the current ensemble and selects additions according to the enlarged ensemble's performance within IG-DOE.
Together, SCOE and IG-DOE couple automatic ensemble construction with coordinated execution.
To the best of our knowledge, this is the first attempt to use LLMs to evolve an ordered destruction-operator ensemble.
We evaluate the evolved IG-DOE on the most challenging public PFSP benchmark (i.e., \textit{VRF-hard-large}~\cite{VRF}) and on a new industrial-data-derived benchmark set, which is constructed from real-world production data collected at an instrument-manufacturing workshop.
The results demonstrate that \mbox{IG-DOE} outperforms the strongest existing methods by a significant margin while retaining cross-scale and cross-benchmark generalization.

The main contributions of this article are summarized below.
\begin{enumerate}
   \item \textbf{A multi-operator IG algorithm for PFSP.}
    We propose IG-DOE, a multi-operator IG algorithm for the PFSP. 
    Instead of relying on a single fixed destruction operator, IG-DOE maintains a DOE and coordinates its use through a stagnation-triggered sequential switching (STSS) mechanism. 
    This design enriches the perturbation behavior of classical IG and improves its ability to escape search stagnation while preserving the efficiency of its construction and local search backbone.

    \item \textbf{An LLM-based framework for automated DOE construction.}
    To reduce reliance on expert-crafted operator design, we introduce the SCOE framework for automatically constructing DOEs. 
    SCOE incrementally expands the ensemble through stagewise evolution, while combining operator-state awareness with task-oriented cooperative evaluation to guide the LLM toward operators that complement the current ensemble.

    \item \textbf{Significant performance improvements on the challenging public benchmark.}
    We evaluate IG-DOE on the \textit{VRF-hard-large} benchmark~\cite{VRF}, which is currently the most challenging benchmark for the PFSP.
    Although the DOE is evolved using only small training instances with $n \in \{100, 200\}$, it generalizes effectively to 180 much larger unseen instances with $n \in \{300,400,500,600,700,800\}$.
    On these large instances, IG-DOE achieves non-trivial improvements over previous state-of-the-art results.
    Specifically, it reduces the average relative percentage deviation (ARPD) from 2.0539\% (QIG) to 1.6874\%, resulting in a 17.8\% improvement.

    \item \textbf{Validation on an industrial-data-derived benchmark.}
    Beyond public benchmarks, we further validate IG-DOE on a new benchmark set derived from actual production logs and processing requirements of an instrument-manufacturing workshop in Shanghai City. 
    Without additional adaptation, IG-DOE (evolved on the small instances in \textit{VRF-hard-large}) delivers the best overall performance among the compared methods, providing further evidence of its generalization ability under realistic manufacturing data distributions.
    To facilitate future research, the source code of IG-DOE and the new industrial-data-derived  benchmark set are anonymously open-sourced: \url{https://github.com/Yufannnnnn/IGDOE-SCOE-PFSP}.

\end{enumerate}

The remainder of this article is organized as follows. 
Section~\ref{sec:related_work} presents the PFSP formulation and reviews related work. 
Section~\ref{sec:online} describes the IG-DOE algorithm and Section~\ref{sec:offline} presents the SCOE framework.
Section~\ref{sec:experiments} presents the experimental evaluation of IG-DOE.
Finally, Section~\ref{sec:conclusion} concludes the article with discussions.         
\section{Problem Definition and Related Work}
\label{sec:related_work}  
This section first formulates the PFSP and then reviews solution methods relevant to this work.
We then discuss multi-operator search strategies for shop scheduling and existing LLM-AHD methods.

\subsection{Problem Definition of the PFSP}
The PFSP is a fundamental shop scheduling problem that was first studied by Johnson in 1954~\cite{PFSPNP1:johnson1954optimal}.
It involves scheduling a set of $n$ jobs $N=\{1,\dots,n\}$ on a set of $m$ machines $M=\{1,\dots,m\}$ according to a fixed processing order $1\to2\to\cdots\to m$. The processing times are given by a matrix $P=[p_{i,j}]$, where $p_{i,j}$ denotes the processing time of job $j$ on machine $i$. A feasible solution is represented by a permutation $\pi=(\pi_1,\dots,\pi_n)$ of job indices.
The job at position $j$ in $\pi$ is the $j$-th job processed on every machine.
For a given $\pi$, the completion time $C_{i,\pi_j}$ of job $\pi_j$ on machine $i$ is calculated using the following recursive equations:
\begin{gather}
C_{1,\pi_1} = p_{1,\pi_1}, \label{eq:completion_first_machine_first_job}\\[2pt]
C_{1,\pi_j} = C_{1,\pi_{j-1}} + p_{1,\pi_j},\ \forall j\ge 2, \label{eq:completion_first_machine}\\[2pt]
C_{i,\pi_1} = C_{i-1,\pi_1} + p_{i,\pi_1},\ \forall i\ge 2, \label{eq:completion_first_job}\\[2pt]
C_{i,\pi_j} = \max\!\bigl\{C_{i-1,\pi_j},\, C_{i,\pi_{j-1}}\bigr\} + p_{i,\pi_j}, \forall i\ge 2,\; j\ge 2. \label{eq:completion_general}
\end{gather}

The goal of the PFSP is to find an optimal permutation $\pi^*$ from the set of all possible permutations $\Omega$ that optimizes a specific objective.
In this work, we focus on the makespan objective, which is the most widely adopted criterion in PFSP research~\cite{Intro_IG_V1:ruiz2007simple, kurdi2020memetic, FernandezViagas2019, QIG:karimi2023learning}.
It is defined as follows:
\begin{equation}
\label{eq:obj_function}
\min_{\pi \in \Omega} \quad C_{\max}(\pi) = C_{m,\pi_n},
\end{equation}
where $C_{m,\pi_n}$ is the completion time of the job at the last position $n$ on the last machine $m$ for a given permutation $\pi$.
Note that although this work focuses on the classical PFSP with makespan minimization, variants with alternative objectives and practical constraints have also been studied in the literature~\cite{de2025systematic, YukselKT24, Jomaa2021_VNS_PM_PFSP, DU2025169}.

\subsection{Solution Methods for the PFSP}
Due to the NP-hard nature of the PFSP~\cite{kan1976machine}, exact algorithms are often computationally prohibitive for medium-to-large instances.
Early studies therefore focused on constructive heuristics, among which the NEH heuristic proposed by Nawaz, Enscore, and Ham remains the most influential representative~\cite{nawaz1983heuristic}.
Beyond constructive heuristics, a wide range of metaheuristics have been developed to improve solution quality within limited computational budgets, including simulated annealing, tabu search, genetic algorithms, ant colony optimization, and memetic algorithms~\cite{OSMAN1989551, RUIZ2005479, kurdi2020memetic}.
More recently, learning-assisted approaches have also been explored, such as Q-learning-guided local search and imitation learning from NEH-based construction rules~\cite{LiLZDZW24, 9953057}.

Among these approaches, the IG algorithm proposed by Ruiz and Stützle remains one of the most effective and widely-adopted methods for the PFSP because of its simple structure and strong empirical performance~\cite{Intro_IG_V1:ruiz2007simple}.
In each iteration, IG applies destruction, reconstruction, local search, and an acceptance mechanism to progressively refine a job permutation.
In practical implementations, the insertion-based operations used in initialization, reconstruction, and local search are commonly accelerated by Taillard's technique~\cite{taillard1990some}, which makes IG particularly attractive as a high-performance backbone for further improvement.
The detailed procedure of the classical IG algorithm is provided in Appendix A of the Supplementary.

Within the IG algorithm, initialization and local search mainly contribute to local improvement, whereas the destruction phase plays the central role in exploration.
Existing improvements to IG have enhanced different components of the algorithm without fundamentally changing this basic structure.
For example, Dubois-Lacoste et al.~\cite{dubois2017iterated} introduced an additional local search on the partial solution after destruction, while Fernandez-Viagas et al.~\cite{FernandezViagas2019} improved tie-breaking across initialization, reconstruction, and local search.
Nevertheless, these methods still rely on a single destruction operator throughout the search.
A notable step toward adapting the destruction phase is QIG~\cite{QIG:karimi2023learning}, which uses Q-learning to adjust the destruction size $d$ during destruction and achieves state-of-the-art performance on the challenging \textit{VRF-hard-large} benchmark~\cite{VRF}.
However, this mechanism remains within a single destruction-operator family, as it changes the perturbation strength but not the removal logic itself.

More broadly, multi-operator search has shown promise in other shop scheduling variants, especially through variable neighborhood search (VNS) and hybrid VNS-based methods~\cite{mladenovic1997variable, BrandimarteFadda2024_RVNS_JITJSP, Jomaa2021_VNS_PM_PFSP, Xu2024_HQPSOVNS_FJSP}.
The idea of this work is related to VNS in the sense that both aim to escape local optima by changing the search behavior during the optimization process.
However, VNS typically switches among predefined neighborhood structures or local search operators, whereas this work focuses on switching among destruction operators within the IG algorithm.

Adaptive large neighborhood search (ALNS) is also closely related to the present work because it follows a destroy-and-repair paradigm and coordinates multiple search operators during optimization~\cite{WINDRASMARA2022}.
Recent studies have applied ALNS to various shop scheduling problems, including distributed reentrant permutation flow shops~\cite{RIFAI2021}, mixed no-wait flow shops~\cite{VerzelloniNagano2022}, dynamic flexible job shops with transportation resources~\cite{LIU2024}, and flexible job shops with multiple resource constraints~\cite{KASAPIDIS2025}.
Similar to the proposed method, these ALNS approaches exploit multiple perturbation behaviors to diversify the search and escape local optima.
However, these approaches generally rely on adaptive mechanisms that update operator-selection probabilities according to historical search performance.
In contrast, our method coordinates destruction operators through stagnation-driven sequential activation rather than online probability updates.

In addition to the differences in operator coordination discussed above, a more fundamental distinction lies in how the operator sets are constructed.
The VNS and ALNS approaches discussed above typically use predefined, manually designed operator sets, whereas our proposed SCOE framework aims to automatically construct high-performing destruction-operator ensembles for IG.
In this work, we focus on the PFSP, although the underlying idea may also be relevant to other IG-based shop scheduling variants.

\subsection{Automatic Heuristic Design}

The automatic discovery of high-performance heuristics has long been an important pursuit in optimization. Genetic programming (GP) has been one of the most widely studied techniques for automated heuristic design, particularly within generative hyper-heuristic research~\cite{zhang2023survey,hh2024}. In production scheduling, GP has been extensively used to evolve dispatching rules and priority functions, with representative studies examining rule representations and GP-based hyper-heuristics for job-shop and dynamic scheduling problems~\cite{Nguyen2013,Fan2021}. Ensemble-based GP has further extended this line of research by combining multiple evolved scheduling heuristics and exploiting their diversity and complementarity in dynamic job-shop scheduling~\cite{PARK2018,XuMZZ24a}. Recent approaches have also investigated the joint evolution of individual heuristics and ensembles, as well as richer program representations for evolved scheduling rules~\cite{XuMZZ24a,HuangMZZ25}. Beyond job-shop scheduling, GP has been applied directly to the PFSP, for example, to automatically discover variants of the NEH construction heuristic~\cite{GPNEH2011}. 
Despite these advances, conventional GP-based heuristic design typically relies on predefined primitives and program representations, which could be difficult to specify appropriately in practice~\cite{EOH:liu2024evolution,langdon2002foundations,zhang2023survey}.
The GP-based scheduling studies reviewed above mainly evolve dispatching or priority rules, with ensemble members typically cooperating by aggregating their outputs at scheduling decision points.

More recently, advances in LLMs have enabled a new form of automatic heuristic design (AHD), in which LLMs generate and refine executable heuristic code within evolutionary frameworks~\cite{liu2026systematic, WuWWFT25,zhang2026llm,lv2026ahd}.
Representative methods such as FunSearch~\cite{FunSearch:romera2024mathematical}, EoH~\cite{EOH:liu2024evolution}, and ReEvo~\cite{ReEvo:ye2024reevo} have established the basic LLM-AHD paradigm.
In these frameworks, LLMs iteratively generate heuristic candidates, while evolutionary selection, reflection, or other search mechanisms retain and refine high-performing ones.
Subsequent studies have extended this paradigm along two main directions. 
One direction is problem-side expansion, where LLM-AHD is adapted to increasingly diverse optimization settings, including Bayesian optimization, continuous optimization, lot-streaming scheduling, and multi-objective optimization~\cite{aglietti2024funbo, SteinB25, li2025llmassisted, huang2025autonomous, Yao00L0025}. 
Alongside these applications, recent studies have strengthened the heuristic discovery process through Monte Carlo tree search~\cite{zheng2025monte}, feature-based niching~\cite{hu2026partition}, and adaptive prompting with accumulated search experience~\cite{zhong2026hifo, wu2026refineevo}.
Other studies investigate game-theoretic co-evolution of solver and instance-generator programs~\cite{ke2026gametheoretic}.
These developments expand how heuristics are generated and evaluated during automatic design.

Despite this rapid progress, most existing LLM-AHD studies still focus on designing a single heuristic component at a time.
More recently, Liu et al.~\cite{liu2025eohs} proposed EoH-S to evolve complementary heuristic sets, marking an important step toward LLM-based multi-heuristic design.
However, such methods still target heuristic sets that are executed under static parallelism, rather than coordinated within a single search trajectory.
This setting differs fundamentally from the problem considered in this work, where multiple destruction operators in the DOE must cooperate sequentially inside one IG algorithm run.
Consequently, the automatic construction of DOEs remains largely open.

\section{The IG-DOE Algorithm}
\label{sec:online}

To avoid confusion, we distinguish between offline DOE construction and online optimization process.
In the offline phase, SCOE constructs an ordered DOE.
In the online phase, IG-DOE takes the constructed DOE and uses it to solve PFSP instances.
This section focuses on the online phase and assumes that a DOE $\mathcal{O}^*=\{O_0,\ldots,O_{K-1}\}$ is already available.
Section~\ref{sec:offline} then explains the offline DOE construction process.

Given $\mathcal{O}^*$, the proposed IG-DOE algorithm follows the main loop of classical IG~\cite{Intro_IG_V1:ruiz2007simple}, but replaces the single fixed destruction operator with a set of heterogeneous destruction operators.
During the search, IG-DOE monitors stagnation and switches the active operator when the current operator no longer improves the global best solution.

\subsection{Overall Framework}
The overall procedure of IG-DOE is summarized in Algorithm~\ref{alg:ig_doe}.
The algorithm iteratively applies destruction (line~10), construction (line~11), local search (line~12), and acceptance (line~13) to improve a solution until the given runtime limit is reached (line~8).
The initial solution is generated by the NEH heuristic (line~3), which is widely used for constructing high-quality PFSP solutions~\cite{FernandezViagas2019,fernandez2020generalised}.
For local search, IG-DOE uses the same insertion neighborhood as classical IG.
For completeness, the pseudocode of NEH~\cite{nawaz1983heuristic} and the insertion-based local search~\cite{Intro_IG_V1:ruiz2007simple} is provided in Appendix B of the Supplementary.
For solution acceptance, IG-DOE adopts the Metropolis criterion~\cite{metropolis1953equation}, same as classical IG.
Specifically, a non-improving candidate $\hat{\pi}$ is accepted with probability $\exp((C_{\max}(\pi)-C_{\max}(\hat{\pi}))/T)$, where $T$ is computed as
\begin{equation}
\label{eq:accept}
T = t \cdot \frac{\sum_{i=1}^{n}\sum_{j=1}^{m} p_{ij}}{n \cdot m \cdot 10}.
\end{equation}

\begin{algorithm}[tbp]
\caption{IG-DOE Algorithm for PFSP}
\label{alg:ig_doe}
\begin{algorithmic}[1]
\STATE \textbf{Input:} Processing times $P$, jobs $n$, machines $m$, ordered DOE $\mathcal{O}^* = (O_0, \dots, O_{K-1})$ (size $K$), temperature factor $t$, stagnation threshold $\tau$.
\STATE \textbf{Output:} Global best permutation $\pi^*$, best makespan $C_{\max}^*$.
\STATE $\pi \leftarrow \text{NEH\_Initialization}()$
\STATE $\pi \leftarrow \text{LocalSearch}(\pi)$
\STATE $\pi^* \leftarrow \pi$, $C_{\max}^* \leftarrow C_{\max}(\pi)$
\STATE $k \leftarrow 0$
\STATE $cnt \leftarrow 0$
\WHILE{termination criterion not met}
    \STATE $O_{curr} \leftarrow O_k$
    \STATE $(\pi_D, \pi_R) \leftarrow \text{Destruction}(\pi, O_{curr})$
    \STATE $\pi' \leftarrow \text{Construction}(\pi_D, \pi_R)$
    \STATE $\pi' \leftarrow \text{LocalSearch}(\pi')$
    \STATE $\pi \leftarrow \text{Acceptance}(\pi', \pi, t)$
    \IF{$C_{\max}(\pi) < C_{\max}^*$}
        \STATE $\pi^* \leftarrow \pi$; $C_{\max}^* \leftarrow C_{\max}(\pi)$
        \STATE $cnt \leftarrow 0$
    \ELSE
        \STATE $cnt \leftarrow cnt + 1$
        \IF{$cnt \ge \tau$}
            \STATE $k \leftarrow (k + 1) \mod K$ \quad \textit{// Switch operator}
            \STATE $cnt \leftarrow 0$
        \ENDIF
    \ENDIF
\ENDWHILE
\STATE \textbf{return} $\pi^*, C_{\max}^*$
\end{algorithmic}
\end{algorithm}

The main difference between IG-DOE and classical IG is therefore the destruction phase.
Classical IG uses one fixed perturbation operator throughout the search, whereas IG-DOE uses the operator selected from $\mathcal{O}^*$.

IG-DOE employs the stagnation-triggered sequential switching (STSS) mechanism to dynamically switch between destruction operators from the DOE $\mathcal{O}^*=\{O_{0},\ldots,O_{K-1}\}$ according to the current search state.
During the search process, a stagnation counter $cnt$ is maintained to record the number of consecutive iterations in which the global best makespan $C^{*}_{\max}$ is not improved. When $cnt$ reaches a predefined threshold $\tau$, an operator switching action is triggered (line~20).

At each iteration, the algorithm applies the current destruction operator $O_{curr}$ to perturb the incumbent solution~(line~10), followed by construction, local search, and acceptance (lines~11--13).
If no improvement in the global best solution is observed for $\tau$ consecutive iterations, STSS switches to the next operator according to the circular rule $k \leftarrow (k+1)\bmod K$ and resets the stagnation counter (lines~17--21).
This deterministic rule has low control overhead and makes cooperative evaluation in offline DOE construction stable.
The benefit of STSS is most pronounced when the DOE contains heterogeneous perturbation behaviors.
This motivates the SCOE framework in Section~\ref{sec:offline}, which is designed to construct such an ensemble automatically.

\section{The SCOE Framework for Automated DOE Construction}
\label{sec:offline} 
Section~\ref{sec:online} describes the IG-DOE algorithm under the assumption that a DOE is already available.
This section explains how such a DOE is constructed offline.
The proposed stagewise cooperative operator ensemble evolution (SCOE) framework builds the DOE one operator at a time.
At each stage, it uses an LLM-driven evolutionary engine, called Co-ReEvo, to generate candidate destruction operators and evaluates each candidate by how well it cooperates with the operators already existing in the DOE.

\subsection{Design Objective of the DOE}
Let $\Omega$ denote the space of executable destruction-operator codes that can be generated by an LLM.
The goal is not to find a single strong operator, but to construct an ordered ensemble $\mathcal{O}=(O_0,\ldots,O_{K-1})$ that works well with the sequential switching rule in IG-DOE.
Since enumerating all possible PFSP instances is impractical, SCOE assumes that a finite training set $\mathcal{I}_{train}=\{I_1,\ldots,I_N\}$ has been collected for offline design.
For a given ensemble $\mathcal{O}$, we define its training performance as the mean average relative percentage deviation (ARPD) over this training set:

\begingroup\small
\begin{equation}
\label{eq:fitness}
F(\mathcal{O}) =
\frac{
	\sum_{I_i \in \mathcal{I}_{train}}
	\left(
	\frac{
		\frac{1}{R}\sum_{r=1}^{R} M(\mathcal{O}, I_i, r) - UB(I_i)
	}{
		UB(I_i)
	}
	\times 100\%
	\right)
}{
	|\mathcal{I}_{train}|
},
\end{equation}\endgroup
where $M(\mathcal{O}, I_i, r)$ is the makespan obtained by IG-DOE using ensemble $\mathcal{O}$ on instance $I_i$ in the $r$-th run, and $UB(I_i)$ is the best-known solution for that instance.
The inner average is taken over the $R$ repeated runs on the same instance, while the outer average is taken over all training instances.
A smaller $F(\mathcal{O})$ indicates better training performance.

The offline DOE design problem can then be written as
\begin{equation}
\label{eq:offline_obj}
\mathcal{O}^*
= \underset{\mathcal{O}: O_k\in\Omega,\ |\mathcal{O}|\le K_{\max}}{\arg\min}
\ F(\mathcal{O}).
\end{equation}
Directly solving Eq.~\eqref{eq:offline_obj} is difficult because it requires searching over many combinations of executable operator codes.
SCOE therefore decomposes the problem into a sequence of smaller tasks, each of which adds one new operator to the current ensemble.

\subsection{Stagewise Construction of the DOE}
SCOE uses an incremental construction strategy (ICS), which starts from an empty ensemble $\mathcal{O}_0$ and expands the ensemble one operator at a time.
Let $\mathcal{O}_t  \circ c$ denote the ensemble obtained by appending candidate operator $c$ to the current ensemble $\mathcal{O}_t$.
At stage $t$, SCOE aims to select the candidate that gives the best training performance after being appended to $\mathcal{O}_t$:
\begin{equation}
\label{eq:add}
c^* = \underset{c \in \Omega}{\arg\min}\ F(\mathcal{O}_t \circ c).
\end{equation}
For the first stage, no existing operator is available, so SCOE searches for a strong first operator.
For later stages, Eq.~\eqref{eq:add} favors operators that complement the existing ensemble rather than only performing well alone.

\begin{algorithm}[tbp]
	\caption{SCOE for Automatic DOE Construction}
	\label{alg:scoe}
	\begin{algorithmic}[1]
		\REQUIRE Training instances $\mathcal{I}_{train}$, ensemble size upper limit $K_{\max}$
		\REQUIRE Subset size $S$, elite size $N_e$, repeated runs $R_1$ and $R_2$
		\ENSURE Constructed DOE $\mathcal{O}^*$
		\STATE $\mathcal{O}_0 \leftarrow ()$; $F_{prev} \leftarrow \infty$; $t \leftarrow 0$
		\WHILE{$|\mathcal{O}_t| < K_{\max}$}
		\STATE $\mathcal{C}_e \leftarrow \text{Co-ReEvo}(\mathcal{I}_{train}, \mathcal{O}_t, S, R_1, N_e)$
		\FOR{each $c_j \in \mathcal{C}_e$}
		\STATE $F_j \leftarrow \text{CoEval}(c_j, \mathcal{O}_t, \mathcal{I}_{train}, R_2)$
		\ENDFOR
		\STATE $c^* \leftarrow \arg\min_{c_j \in \mathcal{C}_e} F_j$
		\STATE $F_{new} \leftarrow \min_{c_j \in \mathcal{C}_e} F_j$
		\IF{$F_{new} \ge F_{prev}$}
		\STATE \textbf{break}
		\ENDIF
		\STATE $\mathcal{O}_{t+1} \leftarrow \mathcal{O}_t \circ c^*$
		\STATE $F_{prev} \leftarrow F_{new}$; $t \leftarrow t+1$
		\ENDWHILE
		\STATE $\mathcal{O}^* \leftarrow \mathcal{O}_t$
		\RETURN $\mathcal{O}^*$
	\end{algorithmic}
\end{algorithm}

Since $\Omega$ is a large space of executable codes, SCOE does not enumerate candidate operators directly.
Instead, it calls \mbox{Co-ReEvo} to generate a small elite set $\mathcal{C}_e$ of candidate destruction operators.
SCOE then uses CoEval to evaluate each elite candidate.
Here, CoEval denotes a cooperative evaluation routine that measures the training performance of the enlarged ensemble formed by appending $c_j$ to $\mathcal{O}_t$.
Details of CoEval and Co-ReEvo (the candidate-generation procedure) are described in Section~\ref{sec:co-reevo}.

\begin{figure*}[tbp]
	% \raggedleft
    \centering
	\includegraphics[width=0.9\linewidth]{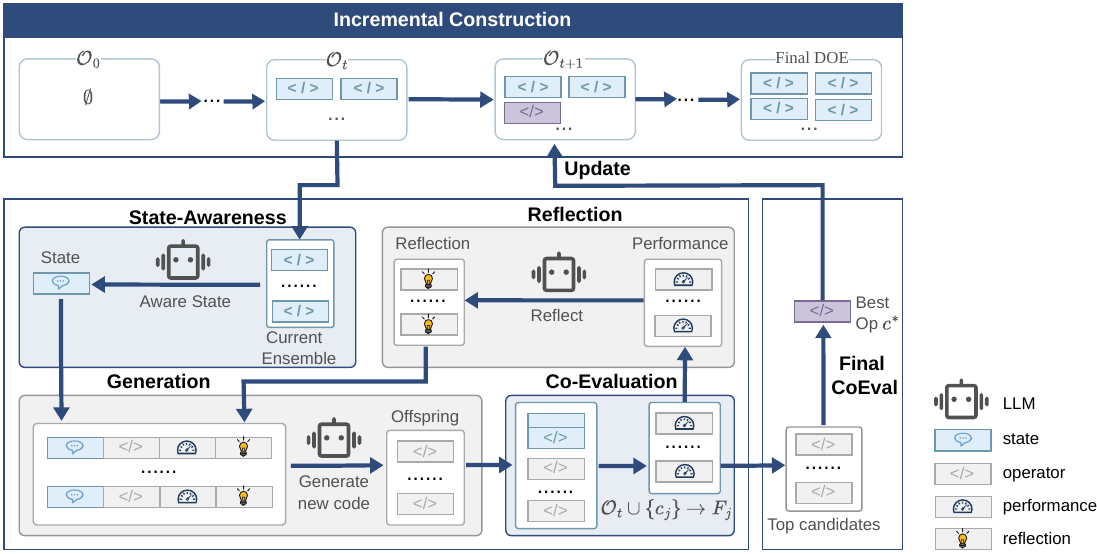} 
	\caption{The proposed SCOE framework for offline DOE construction. SCOE builds an ordered DOE in a stagewise manner. At stage $t$, the current ensemble $\mathcal{O}_t$ is summarized by the state-awareness mechanism and passed to Co-ReEvo. Co-ReEvo generates candidate destruction operators, evaluates them through fast cooperative screening, and returns an elite candidate set. These elite candidates are then evaluated on the full training set after being appended to $\mathcal{O}_t$. The best candidate is accepted to form $\mathcal{O}_{t+1}$ if it improves the ensemble performance, and the process continues until the stopping condition is met.}
	\label{fig:scoe}
\end{figure*}

Algorithm~\ref{alg:scoe} gives the pseudocode of SCOE.
At each stage, Co-ReEvo first produces $\mathcal{C}_e$ using a sampled subset of the training instances (line~3).
SCOE then evaluates these elite candidates on the full training set (lines~4--6) and appends the best one to the DOE if it improves the current ensemble (lines~7--12).
$F_{prev}$ stores the training performance value of the latest accepted ensemble, and $F_{new}$ denotes the best training performance value obtained at the current stage.
If $F_{new}<F_{prev}$, the candidate improves the ensemble and is accepted.
Otherwise, SCOE stops because adding a new operator no longer improves the training performance.
To avoid building an overly large ensemble, SCOE also stops when the ensemble size reaches $K_{\max}$.
Fig.~\ref{fig:scoe} also illustrates the overall offline construction workflow.

\begin{algorithm}[tbp]
	\caption{Co-ReEvo for Candidate Operator Evolution}
	\label{alg:reevo}
	\begin{algorithmic}[1]
		\REQUIRE Training instances $\mathcal{I}_{train}$, current ensemble $\mathcal{O}_t$
		\REQUIRE Subset size $S$, fast repeated runs $R_1$, elite size $N_e$
		\REQUIRE Maximum evaluations $max\_fe$, initial size $N_{init}$, population size $N$
		\ENSURE Elite candidate operators $\mathcal{C}_e$
		\STATE $\mathcal{I}_s \leftarrow \text{RandomSample}(\mathcal{I}_{train}, S)$
		\STATE $State_t \leftarrow \text{ExtractState}(\mathcal{O}_t)$
		\STATE $\mathcal{P} \leftarrow \text{CoEval}(\text{LLMGenerate}(N_{init}, State_t), \mathcal{O}_t, \mathcal{I}_s, R_1)$
		\STATE $\mathcal{A} \leftarrow \mathcal{P}$
		\WHILE{number of evaluations $< max\_fe$}
		\STATE $\mathcal{S}_{pair} \leftarrow \text{SelectPairs}(\mathcal{P}, N)$
		\STATE $R_s \leftarrow \text{ShortTermReflection}(\mathcal{S}_{pair})$
		\STATE $\mathcal{P}_c \leftarrow \text{CoEval}(\text{Crossover}(R_s, N, State_t), \mathcal{O}_t, \mathcal{I}_s, R_1)$
		\STATE $\mathcal{A} \leftarrow \mathcal{A} \cup \mathcal{P}_c$
		\STATE $R_l \leftarrow \text{LongTermReflection}(R_s)$
		\STATE $\mathcal{P}_m \leftarrow \text{CoEval}(\text{Mutate}(R_l, N, State_t), \mathcal{O}_t, \mathcal{I}_s, R_1)$
		\STATE $\mathcal{A} \leftarrow \mathcal{A} \cup \mathcal{P}_m$
		\STATE $\mathcal{P} \leftarrow \text{SelectTop}(\mathcal{P}_c \cup \mathcal{P}_m, N)$
		\ENDWHILE
		\STATE $\mathcal{C}_e \leftarrow \text{SelectTop}(\mathcal{A}, N_e)$
		\RETURN $\mathcal{C}_e$
	\end{algorithmic}
\end{algorithm}
\subsection{Co-ReEvo for Candidate Operator Generation}
\label{sec:co-reevo}
Co-ReEvo is the inner evolutionary engine used at each SCOE stage.
It is built on the reflective evolutionary framework of ReEvo~\cite{ReEvo:ye2024reevo}, where each individual is represented as executable heuristic code and the population is iteratively improved through selection, reflection-guided crossover, and reflection-guided mutation.
Co-ReEvo retains these ReEvo-style evolutionary operations, but adapts them for cooperative operator generation.
The main changes are that the prompt includes the current ensemble state, and the performance of a candidate is computed after appending it to the current ensemble.
Thus, Co-ReEvo does not search for the best standalone operator; it searches for operators that are useful in the current DOE construction stage.

Algorithm~\ref{alg:reevo} summarizes Co-ReEvo.
It first samples a small subset $\mathcal{I}_s$ from the training instances for fast evaluation (line~1).
The LLM then generates an initial population of candidate operators based on the current ensemble state (lines~2--3).
Lines~5--15 follow the ReEvo-style evolutionary loop and are used to produce higher-quality candidates.
Specifically, parent candidates are selected from the current evaluated population, short-term reflection summarizes useful differences between them, crossover generates new candidates, long-term reflection aggregates broader search feedback, mutation introduces additional variations, and the population is updated by selecting the best evaluated candidates.
Detailed descriptions of these ReEvo-style operations are provided in Appendix G of the Supplementary.
Finally, the best candidates found are returned as the elite set $\mathcal{C}_e$ (lines~15--16).

Two modules are central to Co-ReEvo.
The state-awareness mechanism tells the LLM what the current ensemble already contains, while cooperative evaluation (CoEval) tells the search whether a candidate is useful after being added to that ensemble.
They are described below.

\subsubsection{State-Awareness Mechanism}
State-awareness provides the LLM with concise information about the operators already existing in $\mathcal{O}_t$.
At stage $t$, the LLM extracts features from the current operator ensemble $\mathcal{O}_t$ (line~2 of Algorithm~\ref{alg:reevo}), focusing on destruction sizes and specific removal strategies.
These features are summarized as prompt-level state descriptions and inserted into Co-ReEvo prompts before generating new candidates.
The state information contains one concise summary for each existing operator in $\mathcal{O}_t$.
For example, if the current DOE contains three operators, the prompt includes three corresponding summaries; if it contains only one operator, only one summary is inserted.
Table~\ref{tab:state_examples} shows an example extracted during the evolution process, where each row corresponds to one existing operator in the current DOE.
Together, these summaries tell the LLM which perturbation patterns have already been covered, helping it avoid generating operators that are too similar to existing ones and encouraging it to explore missing perturbation patterns.
The complete prompt templates are provided in Appendix D of the Supplementary.

\begin{table}[tbp]
  \centering
  \caption{Example of State Information in One Prompt.}
  \label{tab:state_examples}
  \footnotesize
  \setlength{\tabcolsep}{4pt}
  \renewcommand{\arraystretch}{1.12}
  \begin{tabular}{@{}>{\centering\arraybackslash}p{0.11\columnwidth}
                  >{\raggedright\arraybackslash}p{0.78\columnwidth}@{}}
    \toprule
    Operator & State \\
    \midrule
    $O_0$
    & \textbf{Size:} adaptive 1--4 jobs.\newline
      \textbf{Strategy:} removes jobs with high makespan contribution and adds random perturbation. \\
    \addlinespace[0.25em]
    $O_1$
    & \textbf{Size:} adaptive 2--5 jobs.\newline
      \textbf{Strategy:} combines critical-path jobs, bottleneck-machine jobs, and position-diverse exploration. \\
    \addlinespace[0.25em]
    $O_2$
    & \textbf{Size:} adaptive 2--5 jobs.\newline
      \textbf{Strategy:} selects jobs with high blocking impact and inter-machine dependency complexity using mixed selection. \\
    \bottomrule
  \end{tabular}
\end{table}

\subsubsection{Cooperative Evaluation (CoEval)}
Cooperative evaluation is the key difference between Co-ReEvo and standard single-operator heuristic evolution.
CoEval takes four inputs: a candidate operator $c_j$, the current ensemble $\mathcal{O}_t$, an instance set $\mathcal{I}$, and the number of repeated runs $R$.
It first appends $c_j$ to $\mathcal{O}_t$ to form the enlarged ensemble $\mathcal{O}_t \circ c_j$, and then runs IG-DOE with this enlarged ensemble on each instance in $\mathcal{I}$ for $R$ independent runs.
That is to say, for a candidate $c_j$, the evaluated performance is:
\begin{equation}
\label{eq:candidate_performance}
F_j = F(\mathcal{O}_t \circ c_j).
\end{equation}
This value measures the training performance of IG-DOE after $c_j$ is appended to the current ensemble.
When CoEval receives a set of candidates, it applies the same evaluation procedure to each candidate and returns their performance values.
By comparing such values across candidate operators, CoEval helps identify the candidate that forms the best enlarged ensemble at the current stage.
This evaluation also discourages selecting operators that are strong in isolation but redundant when combined with the current ensemble.

To reduce the offline construction cost, SCOE uses a two-level evaluation scheme.
Inside Co-ReEvo, candidates are evaluated on a randomly sampled subset $\mathcal{I}_s$ with $R_1$ repeated runs, and only the top $N_e$ candidates are kept.
These elite candidates are then evaluated by the outer SCOE procedure on the full training set with $R_2$ repeated runs.
This design keeps the search efficient while making the final operator selection more reliable.

Following~\cite{ReEvo:ye2024reevo}, invalid operators are handled by natural evolutionary filtering.
If a generated operator has a syntax error or fails during execution, CoEval marks it as invalid and excludes it from selection.
Only valid candidates remain available as elites or parents for later generations.

\section{Computational Studies}
\label{sec:experiments}   

This section evaluates IG-DOE and analyzes the design of SCOE.
We first evaluate IG-DOE on the \textit{VRF-hard-large} benchmark~\cite{VRF}, where the DOE is evolved on training instances with $n \in \{100, 200\}$ and tested on larger unseen instances with $n \in \{300, 400, 500, 600, 700, 800\}$.
We then evaluate the method on industrial-data-derived instances to examine generalization under different data distributions.
Next, we examine the main components of SCOE through ablation studies, analyze the actual operators evolved by SCOE and w/o ICS, and evaluate sensitivity to the temperature factor, LLM choice, and stagnation threshold.
Finally, we provide an exploratory study on applying LLM-AHD to different phases of IG.

The experiments aim to answer the following research questions (RQs):
\begin{itemize}
	\item \textit{RQ1}: How does IG-DOE perform on the larger unseen instances of the \textit{VRF-hard-large} benchmark, given that the DOE is constructed from smaller training instances?
	\item \textit{RQ2}: How well does the evolved DOE generalize to industrial-data-derived instances with different data distributions?
	\item \textit{RQ3}: How effectively do the key components of SCOE, namely the incremental construction strategy, state-awareness, and cooperative evaluation, contribute to the quality of the constructed DOE?
    \item \textit{RQ4}: How sensitive is the proposed approach to the temperature factor and stagnation threshold used in IG-DOE and to the LLM API used in offline SCOE construction?
\end{itemize}
To support reproducibility, the source code and the newly constructed industrial-data-derived benchmark are anonymously available at \url{https://github.com/Yufannnnnn/IGDOE-SCOE-PFSP}.

\subsection{Experimental Setup}
\label{subsec:settingsExp}

\subsubsection{Benchmark Sets and Computing Environment}
The \textit{VRF-hard-large}~\cite{VRF} benchmark is used as the public benchmark for the main comparison.
It is regarded as a challenging large-scale benchmark set for PFSP~\cite{QIG:karimi2023learning} and contains 240 instances with the number of jobs $n \ge 100$.
These instances are divided into a training set of 60 smaller instances ($n \in \{100, 200\}$) for constructing the DOE and a testing set of 180 larger instances ($n > 200$) for evaluation.
The testing set is further divided into 18 groups according to the values of $n$ and $m$, with 10 instances in each group.
The industrial-data-derived instances used for RQ2 are described in Section~\ref{subsec:industrialExp}.

In the experiments, all compared algorithms were implemented in Python 3.11 and executed on a Linux server running Ubuntu 22.04.2 LTS, equipped with dual AMD EPYC 9754 128-Core processors and 944 GB of RAM.

\subsubsection{Evaluation Metrics}
The relative percentage deviation (RPD) is used to measure the performance of one independent run on one instance.
For the $r$-th run on a given instance, it is calculated as
\begin{equation}
	\text{RPD}_r = \frac{C_r - UB}{UB} \times 100\%,
	\tag{12}
\end{equation}
where $C_r$ denotes the makespan obtained in the $r$-th run and $UB$ represents the best-known solution for that instance.
Since each instance is independently executed for $R=20$ runs, the average relative percentage deviation (ARPD) of an instance is calculated as
\begin{equation}
	\text{ARPD} = \frac{1}{R}\sum_{r=1}^{R}\text{RPD}_r .
	\tag{13}
\end{equation}
For the industrial-data-derived instances, since no established best-known solutions are available, $UB$ is set to the best observed makespan among all compared algorithms on the same instance.

Following the experimental settings in~\cite{Intro_IG_V1:ruiz2007simple}, all algorithms use the same maximum CPU time limit of $T = n \times (m / 2) \times 0.12$ seconds, where $n$ is the number of jobs and $m$ is the number of machines.

\subsubsection{Compared Methods}
To make the comparison clear, the evaluated methods are grouped into three categories: classical/metaheuristic methods, an LLM-AHD baseline, and IG-DOE variants constructed with different LLMs.

First, to answer RQ1, IG-DOE was compared with four representative PFSP algorithms: IG, QIG, MASC and ALNS.
The classical IG~\cite{Intro_IG_V1:ruiz2007simple} was adopted as the basic baseline, using destruction size $d=4$ and temperature parameter $t=0.4$.
QIG~\cite{QIG:karimi2023learning} was included as a state-of-the-art method on the \textit{VRF-hard-large} benchmark.
Its key parameters were set as follows: local and global improvement weight $\eta=0.3$, temperature scaling factor 0.7, $\epsilon$-greedy initial value $\epsilon=0.8$ with decay rate $\beta=0.996$, learning rate $\alpha=0.6$, discount factor $\gamma=0.8$, episode size $E=6$, and action space $[1, 2, 3]$.
MASC~\cite{kurdi2020memetic} was included as a representative high-performance population-based metaheuristic, with a population size of 100, crossover probability of 0.8, mutation probability of 0.05, and simulated annealing probability of 0.05.
For all these compared algorithms, the parameters were set according to their original papers.
ALNS was included using the open-source implementation of Wouda and Lan~\cite{WoudaLan2023}, with the default basic configuration of its PFSP example.\footnote{\url{https://alns.readthedocs.io/en/latest/examples/permutation_flow_shop_problem.html}}
We additionally adapted the ALNS heuristic of Pascualote Verzelloni and Nagano~\cite{VerzelloniNagano2022} from the mixed no-wait flowshop problem to PFSP. Results for this supplementary baseline, denoted $\text{ALNS}_{\text{VN}}$, are reported in Appendix I of the Supplementary.
We did not include the methods in~\cite{9953057,LiLZDZW24}.
The former was not evaluated on the \textit{VRF-hard-large} benchmark in its original paper and does not provide an open-source implementation, while the latter is an imitation-learning method based on the NEH heuristic and is therefore less directly comparable with metaheuristics that use NEH only for initialization and then perform iterative improvement.

Second, to verify the advantage of our approach over existing LLM-AHD methods, we compared with IG\textsubscript{ReEvo}, an IG variant whose destruction operator is automatically evolved by the ReEvo framework~\cite{ReEvo:ye2024reevo}.
It serves as a strong single-operator LLM-evolved IG baseline.
In its evolution procedure, we used Qwen2.5-Max, with LLM temperature 1.0, population size 10, 30 initial generations, 100 maximum evaluations, crossover rate 1.0, and mutation rate 0.5.

The DOE in IG-DOE was constructed by SCOE using Qwen2.5-Max, the same LLM API used for IG\textsubscript{ReEvo}.
For a fair comparison with the classical IG baseline, IG-DOE and  IG\textsubscript{ReEvo} use the same temperature factor $t=0.4$.
To answer RQ4 regarding the LLM API choice, we additionally construct two IG-DOE variants with DeepSeek-Chat and GPT-4o-mini, which will be analyzed in Section~\ref{subsec:sensitivityExp}.

\begin{table*}[tbp]
	\centering
	\caption{Group-level ARPD results (\%) of the compared algorithms on the 180 {\normalfont\itshape VRF-hard-large} testing instances with $n>200$. 
		Here, $n$ and $m$ denote the numbers of jobs and machines, and ``count'' is the number of instances in each group.
        For each method, ``mean'' and ``best'' report the averages, over the 10 instances in each group, of the mean and minimum RPD values across independent runs on each instance, respectively. 
		Bold values indicate the best result under the corresponding ``mean'' or ``best'' column.}
	\label{tab:comparison_reordered}
	\setlength{\tabcolsep}{4pt}
	\resizebox{0.9\textwidth}{!}{
		\begin{tabular}{ccccccccccccccc}
			\toprule
			\multicolumn{3}{c}{Instance} & \multicolumn{2}{c}{IG} & \multicolumn{2}{c}{MASC} & \multicolumn{2}{c}{QIG} & \multicolumn{2}{c}{$\text{IG}_{\text{ReEvo}}$} & \multicolumn{2}{c}{ALNS} & \multicolumn{2}{c}{IG-DOE} \\
			\cmidrule(lr){1-3} \cmidrule(lr){4-5} \cmidrule(lr){6-7} \cmidrule(lr){8-9} \cmidrule(lr){10-11} \cmidrule(lr){12-13} \cmidrule(lr){14-15}
			$n$ & $m$ & count & mean & best & mean & best & mean & best & mean & best & mean & best & mean & best \\
			\midrule
			300 & 20 & 10 & 2.5576 & 2.3770 & 2.9589 & 2.7607 & 1.7361 & 1.3417 & 1.9494 & 1.5314 & 2.1255 & \textbf{1.0984} & \textbf{1.7227} & 1.3687 \\
			300 & 40 & 10 & 3.5592 & 3.4249 & 4.0775 & 4.0436 & 2.8424 & 2.4063 & 2.6324 & 2.2219 & 3.3306 & 2.6294 & \textbf{2.2807} & \textbf{1.8900} \\
			300 & 60 & 10 & 3.6148 & 3.4303 & 3.9333 & 3.9231 & 2.8889 & 2.5002 & 2.6680 & 2.2256 & 3.3518 & 2.9850 & \textbf{2.1206} & \textbf{1.7270} \\
			400 & 20 & 10 & 1.9790 & 1.8430 & 2.5323 & 2.4161 & 1.4815 & 1.1699 & 1.4714 & 1.1876 & 1.8336 & \textbf{0.8724} & \textbf{1.3168} & 1.0655 \\
			400 & 40 & 10 & 3.2261 & 3.0861 & 3.6605 & 3.6605 & 2.6046 & 2.2751 & 2.5789 & 2.1929 & 3.1209 & 2.7422 & \textbf{2.1696} & \textbf{1.8473} \\
			400 & 60 & 10 & 3.2183 & 3.0944 & 3.5594 & 3.5535 & 2.7223 & 2.4357 & 2.4205 & 2.0233 & 3.0798 & 2.4313 & \textbf{1.9854} & \textbf{1.6078} \\
			500 & 20 & 10 & 1.8119 & 1.6766 & 2.2576 & 2.1153 & 1.2364 & 0.9781 & 1.3966 & 1.1765 & 1.5885 & 1.1586 & \textbf{1.1711} & \textbf{0.9440} \\
			500 & 40 & 10 & 2.9230 & 2.8302 & 3.1969 & 3.1824 & 2.5246 & 2.2372 & 2.3552 & 1.9842 & 2.8041 & 2.5832 & \textbf{2.0354} & \textbf{1.7350} \\
			500 & 60 & 10 & 3.0161 & 2.9145 & 3.1151 & 3.1151 & 2.5176 & 2.2207 & 2.3322 & 2.0118 & 2.9047 & 2.2101 & \textbf{1.9459} & \textbf{1.6029} \\
			600 & 20 & 10 & 1.4533 & 1.3388 & 1.5704 & 1.5704 & 1.0136 & 0.7745 & 1.0878 & 0.8650 & 1.2158 & 1.0393 & \textbf{0.9313} & \textbf{0.7112} \\
			600 & 40 & 10 & 2.7907 & 2.7198 & 3.1346 & 3.1337 & 2.4078 & 2.1785 & 2.2996 & 1.9842 & 2.7239 & 2.3815 & \textbf{2.0021} & \textbf{1.7245} \\
			600 & 60 & 10 & 2.7989 & 2.7373 & 2.9344 & 2.9344 & 2.4607 & 2.2473 & 2.2535 & 2.0105 & 2.7810 & 2.3276 & \textbf{1.8826} & \textbf{1.6301} \\
			700 & 20 & 10 & 1.2147 & 1.1337 & 1.3994 & 1.3863 & 0.8809 & 0.7282 & 0.9530 & 0.7664 & 1.0836 & 0.8636 & \textbf{0.8077} & \textbf{0.6494} \\
			700 & 40 & 10 & 2.4959 & 2.4185 & 2.7725 & 2.7725 & 2.2573 & 2.0197 & 2.1161 & 1.8648 & 2.4087 & 1.8295 & \textbf{1.8570} & \textbf{1.6178} \\
			700 & 60 & 10 & 2.6034 & 2.5430 & 2.7511 & 2.7511 & 2.3837 & 2.2074 & 2.0617 & 1.7907 & 2.5882 & 2.3934 & \textbf{1.7917} & \textbf{1.5620} \\
			800 & 20 & 10 & 1.0646 & 0.9801 & 1.2298 & 1.2162 & 0.7473 & 0.5848 & 0.8070 & 0.6381 & 0.8874 & 0.7486 & \textbf{0.7034} & \textbf{0.5562} \\
			800 & 40 & 10 & 2.4571 & 2.3772 & 2.4588 & 2.4588 & 2.0086 & 1.8201 & 2.0689 & 1.7986 & 2.3521 & 2.0641 & \textbf{1.8460} & \textbf{1.6678} \\
			800 & 60 & 10 & 2.5866 & 2.5267 & 2.7068 & 2.7068 & 2.2554 & 2.1153 & 2.1081 & 1.8311 & 2.6547 & 2.3106 & \textbf{1.8037} & \textbf{1.5903} \\
			\midrule
			Average & -- & -- & 2.5206 & 2.4140 & 2.7916 & 2.7611 & 2.0539 & 1.7912 & 1.9756 & 1.6725 & 2.3797 & 1.9260 & \textbf{1.6874} & \textbf{1.4165} \\
			\bottomrule
		\end{tabular}
	}
\end{table*}

\subsubsection{Parameter Settings of IG-DOE and SCOE}
For IG-DOE, the main parameter is the stagnation threshold $\tau$ in the STSS mechanism, which determines when the active destruction operator is switched.
To account for the effect of instance size on stagnation behavior, we set $\tau$ according to the number of jobs:
\begin{equation}
	\tau = \lfloor C_\tau/\sqrt{n} \rfloor, \tag{7}
\end{equation}
where $n$ denotes the number of jobs and $C_\tau$ is a scaling coefficient.
This scale-sensitive rule is based on our empirical observation that stagnation length is mainly influenced by the job dimension $n$, whereas the machine count $m$ does not show a regular correlation with stagnation length.
It helps avoid overly frequent switching on small instances and delayed switching on large ones.
In this study, $C_\tau$ is set to 1000, which makes $\tau$ approximate the 99th percentile of the observed stagnation periods across instances of different sizes.
The detailed procedure used to derive this setting is provided in Appendix C of the Supplementary.

For offline DOE construction, SCOE uses $K_{\max}=6$ as the maximum ensemble size.
In the two-level cooperative evaluation, candidate operators are first evaluated on a randomly sampled subset of $S=10$ training instances with $R_1=5$ repeated runs per instance, and the top $N_e=10$ operators are retained.
These selected operators are then re-evaluated on the full training set ($|\mathcal{I}_{train}|=60$) with $R_2=10$ runs per instance.
For constructing the DOE, SCOE stopped before reaching $K_{\max}$ because adding the next operator no longer improved the training performance.
It therefore produced a 4-operator DOE, with a total offline construction time of roughly 23 hours on our reference machine.
Details of the parameters and the destruction operators in the evolved DOE are provided in Appendices E and F of the Supplementary, respectively.

\subsection{Performance and Cross-Scale Generalization on \textit{VRF-Hard-Large}}
\label{subsec:comparisonExp}

Table~\ref{tab:comparison_reordered} reports the ARPD results on the 180 larger unseen \textit{VRF-hard-large} testing instances with \(n>200\), where all methods are evaluated under the same CPU-time limit described in Section~\ref{subsec:settingsExp}.
Since the DOE of IG-DOE is evolved only on smaller training instances with $n \in \{100,200\}$, these results evaluate both optimization performance and cross-scale generalization.
                 
\begin{figure*}[t]
	\centering
	\includegraphics[width=0.9\textwidth]{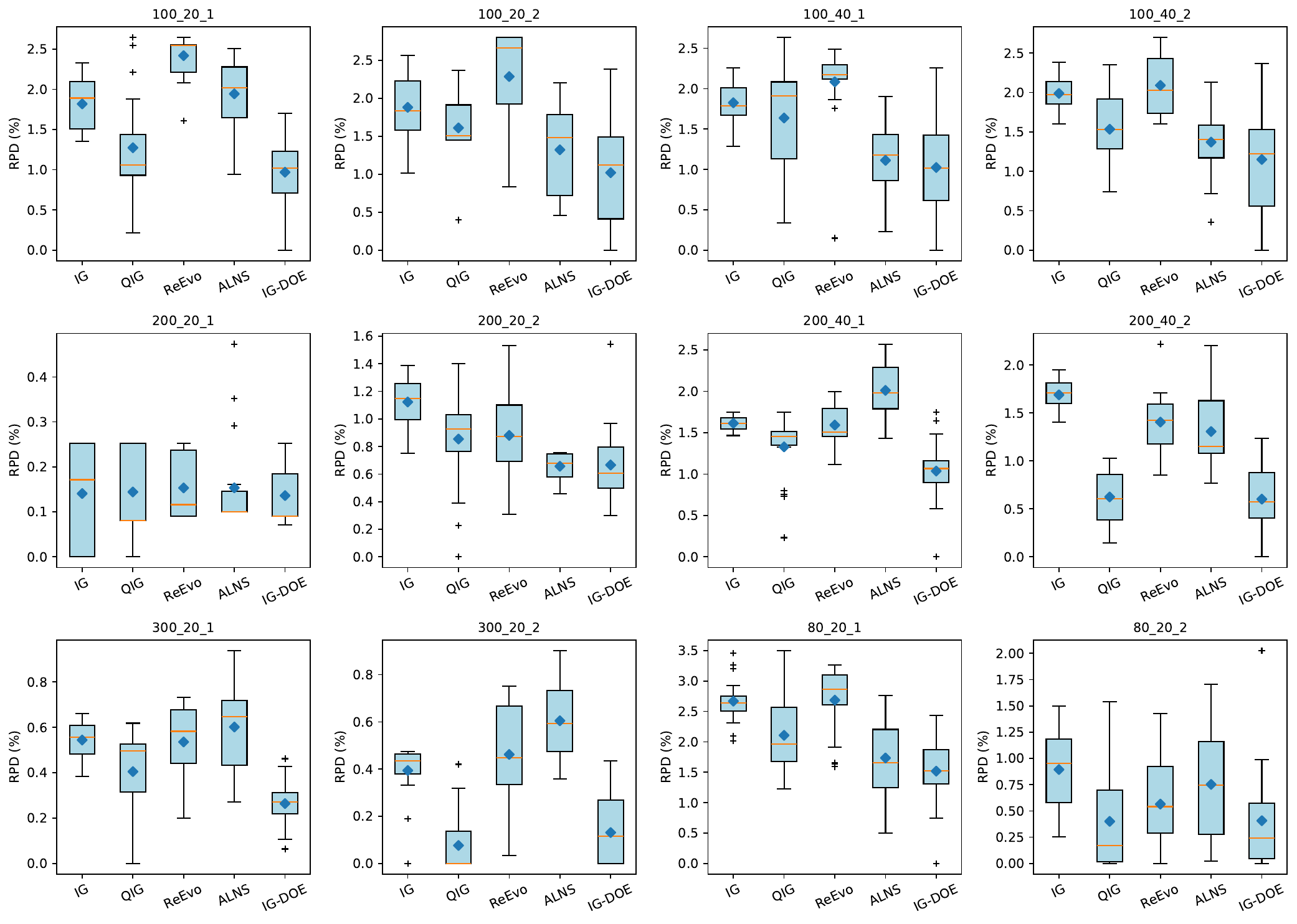}
	\caption{Distributions of RPD values, in terms of boxplots, over 20 independent runs on each of the 12 industrial-data-derived instances. The orange horizontal lines and blue diamonds indicate the median and mean RPD values, respectively.}
	\label{fig:real-data}
\end{figure*}

\begin{table}[bp]
	\centering
	\caption{Positive rank sums of the Wilcoxon signed-rank test on the 18 test groups.}
	\label{tab:wilcoxon_grid}
	\resizebox{\columnwidth}{!}{
		\begin{tabular}{|l|c|c|c|c|c|c|}
			\hline
			& (1) & (2) & (3) & (4) & (5) & (6) \\ \hline
			IG (1) & \text{--} & 171.0 & 0.0 & 0.0 & 4.0 & 0.0 \\ \hline
			MASC (2) & 0.0 & \text{--} & 0.0 & 0.0 & 0.0 & 0.0 \\ \hline
			QIG (3) & 171.0 & 171.0 & \text{--} & 43.0 & 171.0 & 0.0 \\ \hline
			$\text{IG}_{\text{ReEvo}}$ (4) & 171.0 & 171.0 & 128.0 & \text{--} & 171.0 & 0.0 \\ \hline
			ALNS (5) & 167.0 & 171.0 & 0.0 & 0.0 & \text{--} & 0.0 \\ \hline
			IG-DOE (6) & 171.0 & 171.0 & 171.0 & 171.0 & 171.0 & \text{--} \\ \hline
		\end{tabular}
	}
\end{table}

Overall, IG-DOE outperforms all comparison baselines in terms of average ARPD.
Specifically, IG-DOE achieves an overall average ARPD of 1.6874\%, which is lower than the baseline IG value of 2.5206\% and the MASC value of 2.7916\%.
Compared with QIG, the strongest existing IG-based baseline on this benchmark, IG-DOE reduces the overall average ARPD from 2.0539\% to 1.6874\%, corresponding to a relative reduction of 17.8\%.
Compared to IG\textsubscript{ReEvo}, which evolves a single destruction operator using ReEvo, IG-DOE reduces the overall average ARPD by approximately 14.6\%.
Compared with ALNS, IG-DOE reduces the overall mean ARPD by 29.1\% and achieves lower values in 16 of the 18 groups in the best column.
These results support the benefit of constructing a cooperative DOE rather than relying on either destruction-size adaptation or a single LLM-evolved operator.
Since the DOE is evolved only on smaller training instances, its strong performance on the larger unseen instances also confirms its cross-scale generalization ability, providing a positive answer to RQ1.

To further examine whether the advantage is consistent across instance scales, Table~\ref{tab:wilcoxon_grid} reports the Wilcoxon signed-rank statistics computed from the paired group-level mean ARPD values over the 18 test groups.
Each entry corresponds to the positive rank-sum statistic $R^{+}$ of the row method against the column method.
With $G=18$ instance groups, the theoretical maximum is $G(G+1)/2=171$.
The results show that IG-DOE consistently outperforms all five baselines across the test groups.
These statistical results further confirm that the evolved DOE maintains its advantage across different larger instance scales.

\subsection{Comparison on Industrial-Data-Derived Instances}
\label{subsec:industrialExp}

To address RQ2, all algorithms were tested on an industrial-data-derived benchmark constructed from time-study records collected from a measurement-instrument manufacturing workshop in Shanghai, China.
The construction uses recorded step times and aggregated process durations to incorporate empirical information about processing-time differences across production stages.
This differs from the generation of \textit{VRF-hard-large}, where processing times are sampled uniformly from $[1,99]$ and instances are subsequently selected based on difficulty~\cite{VRF}.
The resulting benchmark contains 12 instances covering six job--machine scales, namely 80\_20, 100\_20, 100\_40, 200\_20, 200\_40, and 300\_20, with two instances per scale.
These instances have more right-skewed processing-time distributions, greater relative dispersion, and more uneven cumulative machine workloads than \textit{VRF-hard-large}; detailed statistics are provided in Appendix H of the Supplementary.

Importantly, the DOE used by IG-DOE is still the one evolved from the smaller \textit{VRF-hard-large} training instances with $n \in \{100,200\}$.
No additional DOE construction or parameter adaptation was performed for the industrial-data-derived instances.
Each algorithm is independently executed 20 times on every instance.
Since no established best-known solutions are available for these new instances, ARPD is calculated using the best observed makespan among all compared algorithms on the same instance as the reference value.

The experimental results are shown in Fig.~\ref{fig:real-data}, where each boxplot presents the distribution of RPD values over 20 independent runs on each instance.
MASC is omitted from Fig.~\ref{fig:real-data} for readability due to its poorer performance; the complete boxplot and detailed numerical results are provided in Appendix H of the Supplementary.
IG-DOE achieves a lower mean RPD than QIG on 10 of the 12 instances, with the exceptions of 300\_20\_2 and 80\_20\_2. On 200\_20\_1, its mean RPD is only slightly lower, while QIG obtains better median and best-run results.
These results indicate that the evolved DOE can transfer from synthetic public benchmark instances to data distributions shaped by industrial production, thereby providing evidence of cross-distribution generalization and answering RQ2.

\subsection{Ablation Study}
\label{subsec:ablationExp}
To address RQ3, we examine the individual contributions of the three main design choices in SCOE: incremental construction strategy, state-awareness, and cooperative evaluation.
These components correspond to the stagewise evolution, operator-state awareness, and task-oriented cooperative evaluation introduced in Section~\ref{sec:offline}.
All variants use the same parameters as in Section~\ref{subsec:comparisonExp}, and the temperature factor is fixed at $t=0.4$ to align with the main comparison.

We compare the complete SCOE with three ablation variants, denoted as w/o ICS, w/o State-Awareness, and w/o CoEval.
Specifically, w/o ICS removes the stagewise construction process and uses a ``one-shot generation'' mode, where the LLM generates a complete ensemble with 4 operators simultaneously.
w/o State-Awareness removes the descriptions of existing operator logic from the prompt and retains only the basic task instructions.
w/o CoEval replaces cooperative evaluation with an independent elite-selection strategy.
In this variant, each candidate is evaluated as a singleton ensemble, rather than after being appended to the current ensemble.
For fairness, the ensemble size of w/o ICS is fixed at $K=4$, which matches the size generated by the complete SCOE.
The w/o State-Awareness and w/o CoEval variants keep the original adaptive stopping rule, so their final ensemble sizes are still determined by both the maximum ensemble size and whether adding a new operator improves the training performance.

Table~\ref{tab:ablation} summarizes the ARPD performance of each variant on the testing set.
The complete SCOE achieves the best overall performance, with the lowest mean ARPD in the mean column (1.6874\%) and in the best column (1.4165\%).

\begin{table}[tbp]
  \centering
  \caption{Group-level ARPD results of the ablation study on the key components in SCOE. For each variant, ``mean'' and ``best'' report the averages, over the 10 instances in each group, of the mean and minimum RPD values across independent runs on each instance, respectively.}
  \label{tab:ablation}
  \setlength{\tabcolsep}{2pt}
  \resizebox{\columnwidth}{!}{
  \begin{tabular}{ccccccccccc}
    \toprule
    \multicolumn{3}{c}{Instance} & \multicolumn{2}{c}{w/o ICS} & \multicolumn{2}{c}{w/o State-Awareness} & \multicolumn{2}{c}{w/o CoEval} & \multicolumn{2}{c}{SCOE} \\
    \cmidrule(lr){1-3} \cmidrule(lr){4-5} \cmidrule(lr){6-7} \cmidrule(lr){8-9} \cmidrule(lr){10-11}
    $n$ & $m$ & count & mean & best & mean & best & mean & best & mean & best \\
    \midrule
    300 & 20 & 10 & 2.8839 & 2.8501 & 1.7537 & 1.4685 & 1.8700 & 1.5361 & \textbf{1.7227} & \textbf{1.3687} \\
    300 & 40 & 10 & 3.8404 & 3.8116 & 2.2850 & 1.9199 & 2.5618 & 2.1404 & \textbf{2.2807} & \textbf{1.8900} \\
    300 & 60 & 10 & 3.8766 & 3.8211 & 2.1437 & 1.7804 & 2.4842 & 2.0910 & \textbf{2.1206} & \textbf{1.7270} \\
    400 & 20 & 10 & 2.1442 & 2.0794 & 1.3340 & 1.0877 & 1.4466 & 1.1869 & \textbf{1.3168} & \textbf{1.0655} \\ 
    400 & 40 & 10 & 3.4507 & 3.4480 & 2.1944 & 1.8899 & 2.4539 & 2.1076 & \textbf{2.1696} & \textbf{1.8473} \\
    400 & 60 & 10 & 3.3594 & 3.3337 & 1.9969 & 1.6411 & 2.3356 & 2.0240 & \textbf{1.9854} & \textbf{1.6078} \\
    500 & 20 & 10 & 2.0060 & 1.9671 & 1.1942 & \textbf{0.9432} & 1.3285 & 1.0658 & \textbf{1.1711} & 0.9440 \\
    500 & 40 & 10 & 2.9948 & 2.9560 & 2.0484 & 1.7495 & 2.2917 & 2.0251 & \textbf{2.0354} & \textbf{1.7350} \\
    500 & 60 & 10 & 3.1831 & 3.1693 & 1.9874 & 1.6688 & 2.2877 & 2.0082 & \textbf{1.9459} & \textbf{1.6029} \\
    600 & 20 & 10 & 1.6216 & 1.5876 & 0.9477 & 0.7757 & 1.0552 & 0.8742 & \textbf{0.9313} & \textbf{0.7112} \\
    600 & 40 & 10 & 2.8876 & 2.8873 & 2.0307 & 1.7811 & 2.2529 & 1.9652 & \textbf{2.0021} & \textbf{1.7245} \\
    600 & 60 & 10 & 2.8848 & 2.8661 & 1.9438 & 1.6804 & 2.2027 & 1.9535 & \textbf{1.8826} & \textbf{1.6301} \\
    700 & 20 & 10 & 1.2484 & 1.1955 & 0.8263 & 0.6908 & 0.9038 & 0.7585 & \textbf{0.8077} & \textbf{0.6494} \\
    700 & 40 & 10 & 2.5734 & 2.5516 & 1.8654 & 1.6384 & 2.0781 & 1.8374 & \textbf{1.8570} & \textbf{1.6178} \\
    700 & 60 & 10 & 2.7044 & 2.7010 & 1.8126 & 1.5996 & 2.0773 & 1.8992 & \textbf{1.7917} & \textbf{1.5620} \\
    800 & 20 & 10 & 1.2173 & 1.2013 & 0.7380 & 0.5884 & 0.7847 & 0.6552 & \textbf{0.7034} & \textbf{0.5562} \\
    800 & 40 & 10 & 2.5604 & 2.5488 & 1.8739 & \textbf{1.6557} & 2.0704 & 1.8340 & \textbf{1.8460} & 1.6678 \\
    800 & 60 & 10 & 2.6334 & 2.6222 & 1.8305 & 1.6401 & 2.0865 & 1.8934 & \textbf{1.8037} & \textbf{1.5903} \\
    \midrule
    Average & -- & -- & 2.6706 & 2.6443 & 1.7115 & 1.4555 & 1.9206 & 1.6586 & \textbf{1.6874} & \textbf{1.4165} \\
    \bottomrule
  \end{tabular}
  }
\end{table}
Among the three variants, w/o ICS shows the largest performance drop, with its average ARPD in the mean column increasing to 2.6706\%.
This gap indicates that it is difficult for an LLM to design a complete multi-operator ensemble in a single pass.
Without stagewise guidance, one-shot generation makes it much harder to obtain a high-quality operator ensemble.

The effect of state-awareness is shown by w/o State-Awareness, which yields an a average ARPD in the mean column of 1.7115\%.
Although this variant is better than w/o ICS, it is still worse than the complete SCOE.
This suggests that the state information helps the LLM understand what the current ensemble already contains and generate operators that better complement the existing ones.

Furthermore, w/o CoEval obtains an average ARPD of 1.9206\% in the mean column, which is clearly worse than both w/o State-Awareness and the complete SCOE.
This result indicates that selecting operators only by their individual performance is not enough for building an effective ensemble.
When cooperation with the current DOE is ignored, individually strong operators may not combine well with the existing ensemble.
In contrast, the cooperative evaluation used by SCOE evaluates each candidate after it is appended to the current ensemble, thereby guiding the search toward operators that form better enlarged ensembles.

Overall, these results show that the incremental construction strategy, state-awareness, and cooperative evaluation all contribute to the quality of the constructed DOE, providing a positive answer to RQ3.

\begingroup
\subsection{Analysis of the Evolved Operators}
\label{subsec:operatorAnalysis}

The preceding ablation study shows a performance gap between SCOE and w/o ICS. To examine the behavioral differences underlying this gap, Table~\ref{tab:evolved_operator_summary} summarizes the main operations of the two final four-operator ensembles. Complete pseudocode is provided in Appendix F of the Supplementary.

\begin{table}[tbp]
    \centering
    \caption{Evolved destruction operators.}
    \label{tab:evolved_operator_summary}
    \footnotesize
    \setlength{\tabcolsep}{4pt}
    \renewcommand{\arraystretch}{1.12}
    \begin{tabular}{@{}>{\centering\arraybackslash}p{0.15\columnwidth}>{\raggedright\arraybackslash}p{\dimexpr0.85\columnwidth-2\tabcolsep\relax}@{}}
        \toprule
        Operator & Main operation \\
        \midrule
        \multicolumn{2}{@{}l@{}}{\textbf{SCOE}} \\
        \addlinespace[2pt]
        $O_0$ & Remove up to four highest-impact jobs and shuffle their reinsertion order. \\
        $O_1$ & Remove 3--6 highest-impact jobs, with the number sampled uniformly. \\
        $O_2$ & Follow $O_1$, breaking impact ties by longer processing times on the maximum-workload machine. \\
        $O_3$ & Remove 3--6 jobs by randomly selecting either end of the permutation at each step. \\
        \midrule
        \multicolumn{2}{@{}l@{}}{\textbf{w/o ICS}} \\
        \addlinespace[2pt]
        $\widetilde O_0$ & Remove the two lowest-impact jobs. \\
        $\widetilde O_1$ & Remove the four jobs with the longest last-machine processing times. \\
        $\widetilde O_2$ & Remove the six jobs with the longest total processing times. \\
        $\widetilde O_3$ & Behaviorally equivalent to $\widetilde O_0$. \\
        \bottomrule
    \end{tabular}
    \par\smallskip
    \parbox{\columnwidth}{\footnotesize Impact denotes the makespan reduction caused by deleting a job. Complete operator pseudocode is provided in Appendix F of the Supplementary.}
\end{table}

SCOE combines impact-based removal with variations in destruction size, reinsertion order, and positional selection. Operators $O_0$--$O_2$ prioritize jobs with larger deletion impacts. However, $O_0$ uses a makespan-dependent destruction size capped at four and randomizes the reinsertion order, whereas $O_1$ and $O_2$ uniformly sample the destruction size from three to six and retain the selection-ranking order for reinsertion. Operator $O_2$ further distinguishes itself by breaking equal-impact ties using longer processing times on the maximum-workload machine. Operator $O_3$ instead randomly removes jobs from the two ends of the permutation and reinserts them in removal order. Together, these operators offer potentially complementary perturbations through differences in removal strength, reinsertion order, and the jobs selected.

The w/o ICS ensemble exhibits redundancy and more restricted perturbation behaviors. Operators $\widetilde O_0$ and $\widetilde O_3$ are behaviorally equivalent: for the same permutation, both remove the two jobs with the smallest deletion impacts and return identical remaining and removed-job sequences. Operators $\widetilde O_1$ and $\widetilde O_2$ use permutation-independent processing-time scores and therefore tend to select the same job subsets throughout the search. All four operators use fixed destruction sizes and produce the same perturbation when reapplied to the same permutation, potentially limiting their ability to escape local optima.

These behavioral differences provide plausible explanations for the observed performance gap and illustrate the motivation for incremental construction. ICS keeps the existing ensemble fixed and evaluates candidates within that ensemble, focusing feedback on whether an addition improves the combined search performance. In contrast, w/o ICS jointly evolves all four operators, making individual contributions and interactions harder to distinguish from ensemble-level feedback. Thus, ICS structures the design task around successive useful additions, rather than requiring the LLM to refine an entire ensemble at once.

To examine the effect of operator ordering, we compare the original order $(O_0,O_1,O_2,O_3)$ with the reversed order $(O_3,O_2,O_1,O_0)$ and a fixed permuted order $(O_1,O_3,O_0,O_2)$ used to simulate a random ordering. The four evolved operators remain unchanged, and each run starts with the first operator in the specified order, with subsequent switches following STSS. All three configurations use the same time budget and parameter settings, with temperature factor $t=0.4$ and stagnation threshold $\tau=\lfloor 1000/\sqrt{n}\rfloor$. Each configuration is evaluated over 20 independent runs per instance on the 60 training instances and 180 testing instances.

\begin{table}[tbp]
    \centering
    \caption{Mean ARPD (\%) under different operator orders. Results are averaged over 60 training instances and 180 testing instances, with 20 runs per instance. Lower values are better; the best value in each column is highlighted in bold.}
    \label{tab:operator_order}
    \setlength{\tabcolsep}{5pt}
    \begin{tabular}{lccc}
        \toprule
        Order & Operator sequence & Training & Testing \\
        \midrule
        Original & $(O_0,O_1,O_2,O_3)$ & \textbf{2.2689} & \textbf{1.6874} \\
        Reversed & $(O_3,O_2,O_1,O_0)$ & 2.5935 & 1.7357 \\
        Permuted & $(O_1,O_3,O_0,O_2)$ & 2.4764 & 1.8161 \\
        \bottomrule
    \end{tabular}
\end{table}

Table~\ref{tab:operator_order} shows that the original order achieves the lowest mean ARPD on both the training and testing sets, indicating that operator ordering affects the performance of the fixed ensemble. This is consistent with SCOE's stagewise construction, where each new operator is evaluated after being appended to the existing ensemble. The results support retaining the resulting order when applying the DOE to unseen instances.

\endgroup

\subsection{Sensitivity Analysis}
\label{subsec:sensitivityExp}

To address RQ4, we examine sensitivity to the temperature factor $t$, the LLM API used in offline SCOE construction, and the stagnation threshold $\tau$. The temperature-factor and LLM analyses use the 60 training instances ($n\in\{100,200\}$), while the threshold analysis uses the 180 testing instances defined in Section~\ref{subsec:settingsExp}.

\begin{figure}[tbp]
	\centering
	\includegraphics[width=0.9\columnwidth]{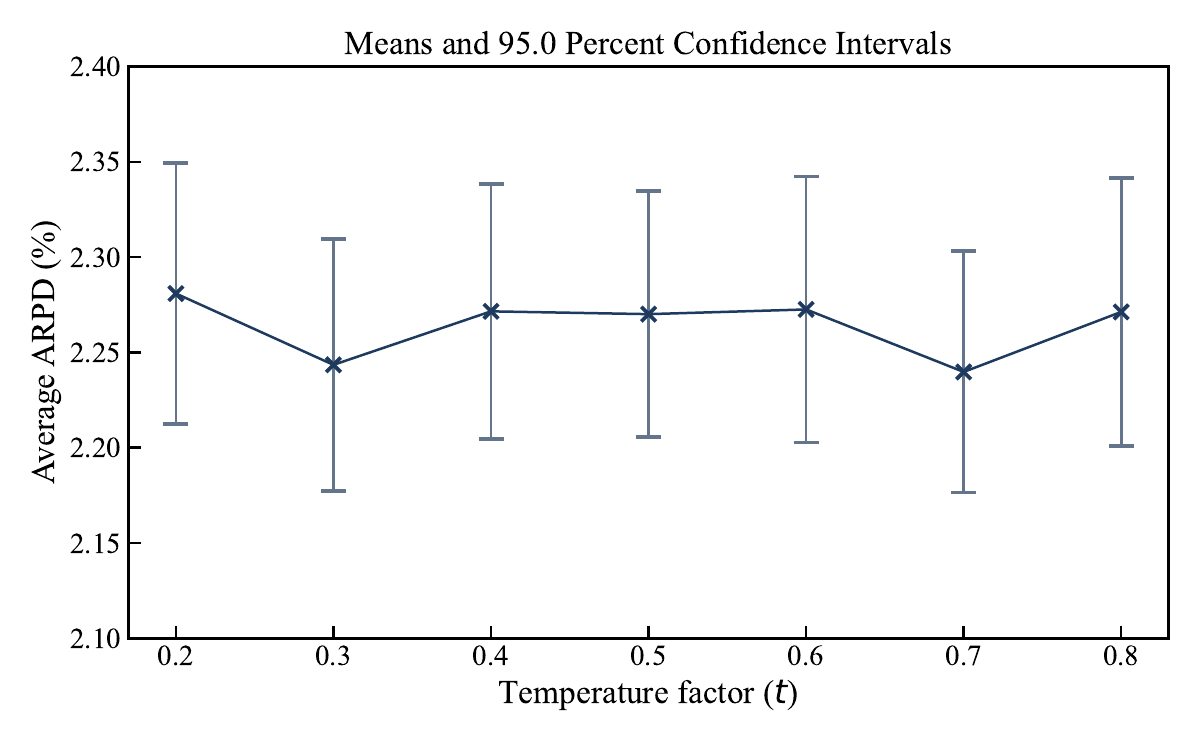}
    \caption{Sensitivity analysis of the temperature factor $t$ on the 60 training instances. Points show the mean ARPD. Error bars show 95\% $t$-based confidence intervals calculated across instance-level mean ARPD values, each averaged over 20 independent runs.}
	\label{fig:temp_analysis}
\end{figure}

First, the temperature factor $t$ in the acceptance criterion controls the trade-off between convergence speed and the ability to escape local optima.
We performed a sensitivity analysis for $t \in [0.2, 0.8]$ with a step size of 0.1.
Fig.~\ref{fig:temp_analysis} reports the mean ARPD for each value of $t$, together with 95\% $t$-based confidence intervals calculated across instance-level mean ARPD values, each averaged over 20 independent runs. The mean ARPD varies only within approximately $[2.24\%,2.28\%]$, corresponding to a maximum difference of approximately 0.04 percentage points. A Friedman test based on the instance-level mean ARPD values, with the 60 training instances as blocks, detected no statistically significant difference among the seven temperature settings ($\chi_F^2=8.5658$, $df=6$, $p>0.05$).
Although $t=0.7$ gives the lowest ARPD in this analysis, the main comparison uses $t=0.4$ to align with the classical IG baseline.
Overall, IG-DOE has limited sensitivity to the temperature factor over the examined range.

Second, we evaluate the sensitivity of SCOE to the choice of LLM API used for offline DOE construction.
We compare the primary choice, Qwen2.5-Max (accessed via the qwen-max API endpoint), with DeepSeek-Chat (deepseek-chat for DeepSeek-V3) and GPT-4o-mini.
The comparative results are summarized in Table~\ref{tab:llm_comparison}.
Although Qwen2.5-Max gives the lowest average ARPD of 2.239775\%, DeepSeek-Chat and GPT-4o-mini also construct effective DOEs, with average ARPD values of 2.341541\% and 2.461430\%, respectively.
This indicates that the LLM API choice affects the final DOE quality, but the two alternative APIs still produce competitive DOEs with about 0.1 and 0.2 percentage-point increases in average ARPD.
Thus, SCOE is not narrowly tied to a single LLM API in these experiments.
This property is useful in practice, because LLMs are black-box systems and their capabilities may change across model versions and service updates.
Furthermore, we record the monetary cost and token consumption of offline DOE construction.
As mentioned in Section~\ref{subsec:settingsExp}, the total cost depends on the number of Co-ReEvo stages executed before the adaptive stopping criterion is met.
For instance, the evolution with Qwen2.5-Max completed 5 stages and produced a 4-operator ensemble, consuming roughly 1 to 2 million tokens with a total cost of \$0.5282.
The full adaptive runs for DeepSeek-Chat and GPT-4o-mini stopped earlier, resulting in final ensemble sizes of 2 and 3 operators, respectively, with total costs of \$0.2427 and \$0.4573.
Since DOE construction is performed only once offline, these API costs are relatively modest for practical use.

\begin{figure}[tbp]
    \centering
    \includegraphics[width=0.85\columnwidth]{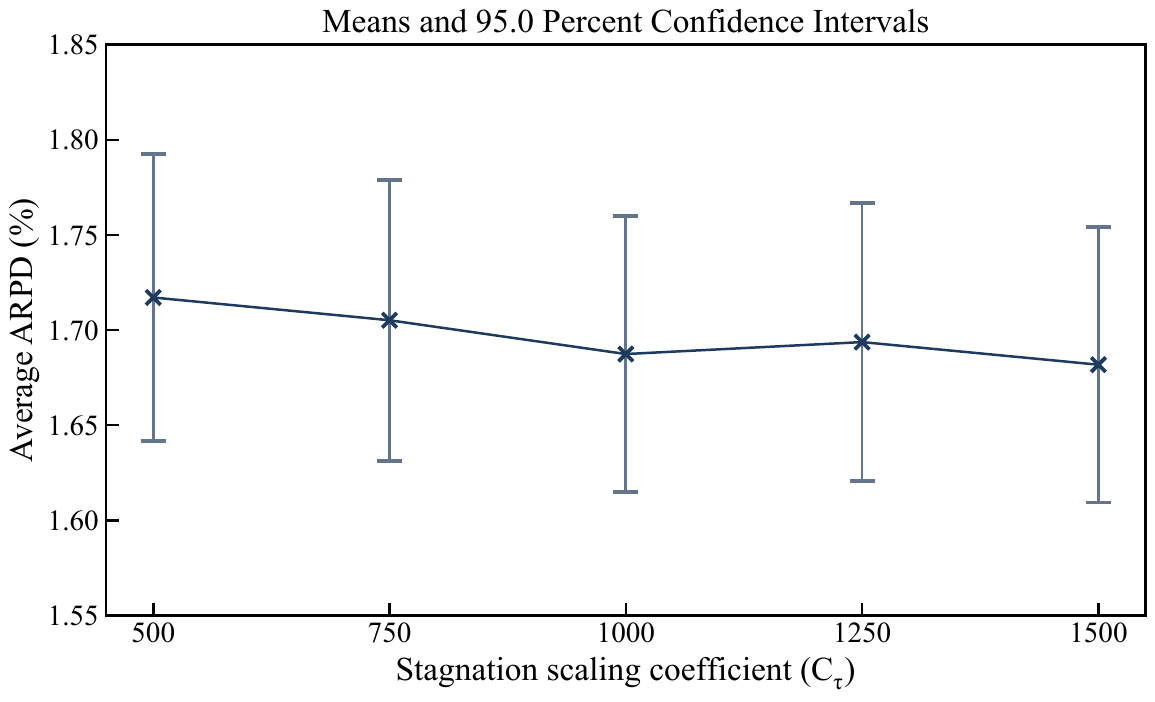}
    \caption{Sensitivity analysis of the stagnation scaling coefficient $C_\tau$ on the 180 testing instances. Points show the mean ARPD. Error bars show 95\% $t$-based confidence intervals calculated across instance-level mean ARPD values, each averaged over 20 runs.}
    \label{fig:stagnation_sensitivity}
\end{figure}

Third, we examine the stagnation threshold by varying $C_\tau\in\{500,750,1000,1250,1500\}$ in $\tau=\lfloor C_\tau/\sqrt{n}\rfloor$. We use the DOE evolved by SCOE with Qwen2.5-Max, keeping its operators and their order fixed. Each setting is evaluated over 20 independent runs per testing instance, with $t=0.4$ and the CPU-time budget specified in Section~\ref{subsec:settingsExp}. As shown in Fig.~\ref{fig:stagnation_sensitivity}, the overall mean ARPD ranges from 1.6818\% to 1.7171\%, a spread of 0.0353 percentage points; the retained default $C_\tau=1000$ yields 1.6874\%. A Friedman test based on the instance-level mean ARPD values, with the 180 testing instances as blocks, detected a statistically significant difference among the five threshold settings ($\chi_F^2=145.3029$, $df=4$, $p<0.001$). Thus, the threshold coefficient affects performance, although the observed changes in overall mean ARPD are small over the examined range under this fixed-DOE configuration.

Overall, these results suggest that the proposed approach exhibits limited variation in average performance across the examined temperature factors, LLM APIs, and stagnation thresholds, providing a positive answer to RQ4.

\begin{table}[tbp]
	\centering
	\begin{threeparttable}
		\caption{Group-level mean ARPD (\%) of IG-DOE with different LLM APIs.}
		\label{tab:llm_comparison}
		\begin{tabular*}{0.9\linewidth}{@{\extracolsep{\fill}}cccccc}
			\toprule
			$n$ & $m$ & count & IG-DOE & IG-DOE2 & IG-DOE3 \\ \midrule
			100 & 20 & 10 & \textbf{2.305774} & 2.650766 & 2.841311 \\
			100 & 40 & 10 & \textbf{2.316241} & 2.413127 & 2.663856 \\
			100 & 60 & 10 & \textbf{2.008681} & 2.065298 & 2.157458 \\
			200 & 20 & 10 & 2.274138 & 2.325456 & \textbf{2.263947} \\
			200 & 40 & 10 & \textbf{2.407371} & 2.450812 & 2.569442 \\
			200 & 60 & 10 & \textbf{2.126443} & 2.143789 & 2.272564 \\ \midrule
			Average & -- & -- & \textbf{2.239775} & 2.341541 & 2.461430 \\ \bottomrule
		\end{tabular*}
		\begin{tablenotes}
			\item IG-DOE (Qwen2.5-Max, \$0.5282), IG-DOE2 (DeepSeek-Chat, \$0.2427), IG-DOE3 (GPT-4o-mini, \$0.4573)
		\end{tablenotes}
	\end{threeparttable}
\end{table}
\begin{table}[tbp]
	\centering
	\footnotesize
	\caption{Group-level mean ARPD (\%) and average number of iterations of IG, ReEvo-LS, and ReEvo-Des (within the same runtime limit).}
	\label{tab:motivation}
	\setlength{\tabcolsep}{0pt}
	\begin{tabular*}{0.9\linewidth}{@{\extracolsep{\fill}}ccccccccc}
		\toprule
		\multicolumn{3}{c}{Instance} & \multicolumn{2}{c}{IG} & \multicolumn{2}{c}{ReEvo-LS} & \multicolumn{2}{c}{ReEvo-Des} \\
		\cmidrule(lr){1-3} \cmidrule(lr){4-5} \cmidrule(lr){6-7} \cmidrule(lr){8-9}
		$n$ & $m$ & count & ARPD & Iter. & ARPD & Iter. & ARPD & Iter. \\
		\midrule
		100 & 20 & 10 & 4.41 & 986 & 3.94 & 11 & 2.71 & 431 \\
		100 & 40 & 10 & 4.29 & 956 & 4.05 & 9 & 2.51 & 394 \\
		100 & 60 & 10 & 3.93 & 979 & 3.69 & 10 & 2.26 & 362 \\
		200 & 20 & 10 & 3.44 & 809 & 3.14 & 7 & 2.35 & 286 \\
		200 & 40 & 10 & 4.17 & 865 & 3.60 & 8 & 2.59 & 286 \\
		200 & 60 & 10 & 3.80 & 791 & 3.36 & 6 & 2.40 & 272 \\
		\midrule
		\multicolumn{3}{c}{Average} & 4.01 & 898 & 3.63 & 9 & 2.47 & 338 \\
		\bottomrule
	\end{tabular*}
\end{table}

\subsection{An Exploratory Study on LLM-AHD in Different IG Phases}
\label{subsec:preExp}

As an additional analysis, we examine what happens if LLM-AHD is applied to different phases of IG.
This study could help reveal the practical trade-off between redesigning the local search phase and redesigning the destruction phase.
Specifically, we used the ReEvo framework~\cite{ReEvo:ye2024reevo} to independently evolve a single component for either the local search phase (ReEvo-LS) or the destruction phase (ReEvo-Des).
Both variants were compared with the baseline IG~\cite{Intro_IG_V1:ruiz2007simple} on the 60 training instances ($n \in \{100, 200\}$) under the same CPU-time limit.

As reported in Table~\ref{tab:motivation}, ReEvo-LS reduces the average ARPD of IG from 4.01\% to 3.63\%, but the average number of iterations drops sharply from 898 to 9.
In contrast, \mbox{ReEvo-Des} achieves a much lower average ARPD of 2.47\% while still completing 338 iterations on average.
The main reason is computational overhead.
The baseline IG uses the Taillard acceleration technique~\cite{taillard1990some} for insertion-neighborhood evaluation, which reduces the time cost of evaluating a move.
However, LLM-generated local search components often contain mixed or nested move patterns, which cannot directly reuse this acceleration and thus require more expensive evaluations.
These observations suggest that, within the IG framework, the destruction phase is a more practical target for LLM-AHD than directly replacing local search.
    
\section{Conclusions}
\label{sec:conclusion}    
This work investigated how to enhance the exploration ability of IG for PFSP without changing its efficient construction and local search backbone.
Instead of relying on a single fixed destruction operator, IG-DOE uses a destruction operator ensemble (DOE) and switches among its operators when search stagnation is detected.
To construct such an ensemble automatically, SCOE evolves destruction operators in a stagewise manner and evaluates each candidate according to its cooperation with the current ensemble.

The experimental results provide several observations.
First, the DOE evolved from smaller \textit{VRF-hard-large} training instances performs well on larger unseen instances, reducing the overall average ARPD from 2.0539\% for QIG to 1.6874\% under the same CPU-time limit.
Second, the same DOE also performs well on industrial-data-derived instances without additional adaptation, suggesting cross-distribution generalization.
Third, the ablation study confirms that incremental construction, state-awareness, and cooperative evaluation all contribute to the quality of the constructed DOE.
The sensitivity analysis further shows that the method is not highly sensitive to the temperature factor or to the tested LLM API choices.

In general, these findings suggest that combining an operator ensemble with SCOE-based automatic construction is an effective way to enhance iterative greedy (IG) algorithms.
Since IG-based algorithms have also shown strong performance on various scheduling variants~\cite{wu2025balance, yu2025self}, the same idea can be naturally extended to these algorithms and problems by adapting the definition of destruction operators to the target scheduling setting and using SCOE to construct heterogeneous operators automatically.
Beyond IG-based scheduling algorithms, this idea may also be applied to other iterative metaheuristic backbones where search stagnation can be alleviated by switching among heterogeneous operators.
Another interesting future direction is to replace deterministic sequential switching with a learned switching policy.
For example, IG-DOE could learn when to switch~\cite{LvWCL25_UNiCS} and could assign higher switching probabilities to operators that are more likely to complement the current search state or escape the current local-optimal region~\cite{PeiTLMY23_LOC_AOS}.

\section*{Acknowledgments}
This work was supported by the National Natural Science Foundation of China under Grant 62502192, and in part by Zhongguancun Academy, (Grant No.s 02012403).

\bibliographystyle{IEEEtran}
\bibliography{references}

\begin{IEEEbiographynophoto}{Rui Xu (Member, IEEE)} received the Ph.D. degree in Management Science and Engineering in 2011 through a joint doctoral program between the University of Science and Technology of China (USTC), Hefei, China, and the University of Tennessee, Knoxville, TN, USA. He is currently a Professor with the School of Business, Hohai University, Nanjing, China. He also serves as the Head of the Department of Management Science and Information Management and the Director of the Business Data Laboratory. His research interests include intelligent optimization and decision-making, evolutionary computation, smart logistics, and supply chain optimization.
\end{IEEEbiographynophoto}

\begin{IEEEbiographynophoto}{Yufan Liao (Student Member, IEEE)} is currently pursuing a graduate degree in Management Science and Engineering at Hohai University, Nanjing, China. Her research interests include combinatorial optimization and foundation models for optimization.
\end{IEEEbiographynophoto}

\begin{IEEEbiographynophoto}{Haoze Lv (Student Member, IEEE)} received the B.Eng. degrees from the Southern University of Science and Technology (SUSTech), Shenzhen, China, in 2023. He is currently working toward a Ph.D. degree in the Department of Computer Science and Engineering at SUSTech. His research interests include automatic algorithm design and machine learning.
\end{IEEEbiographynophoto}

\begin{IEEEbiographynophoto}{Shengcai Liu (Member, IEEE)} received the B.Sc. degree and the Ph.D. degree from the University of Science and Technology of China, Hefei, China in 2014 and 2020, respectively. He is currently an Assistant Professor with the Department of Computer Science and Engineering, Southern University of Science and Technology, Shenzhen, China. He has authored or coauthored more than 30 papers in top-tier refereed international conferences and journals. His research focuses on learning to optimize. He is an Associate Editor of IEEE TRANSACTIONS ON EMERGING TOPICS IN COMPUTATIONAL INTELLIGENCE.
\end{IEEEbiographynophoto}

\begin{IEEEbiographynophoto}{Yi Mei (Senior Member, IEEE)} is a Professor at the School of Engineering and Computer Science, Victoria University of Wellington. His research interests include evolutionary computation and machine learning for combinatorial optimisation, genetic programming, reinforcement learning, automatic algorithm design, explainable AI, multi-objective optimisation, transfer/multitask/lifelong learning and optimisation.
Yi has over 300 fully refereed publications, including the top journals in AI, EC and Operations Research (OR) such as IEEE Transactions on Pattern Analysis and Machine Intelligence (TPAMI), IEEE Transactions on Evolutionary Computation
(TEVC), IEEE Transactions on Cybernetics (TCYB), European Journal of Operational Research, ACM Transactions on Mathematical Software, and top AI/EC conferences (AAAI, ICLR, IJCAI, and GECCO). He won an IEEE Transactions on Evolutionary Computation Outstanding Paper Award 2017, IEEE Computational Intelligence Magazine Outstanding Paper Award 2026, GECCO Best Paper Awards 2022, 2023, 2024, and EuroGP Best Paper Award 2022. He is an Associate Editor of IEEE Transactions on
Evolutionary Computation, IEEE Transactions on Artificial Intelligence,
Computational Intelligence Magazine, and IEEE Computational Intelligence Letter. He is the Chair of the IEEE CIS Evolutionary Computation Technical Committee and the Chair of IEEE New Zealand Council. He is a Fellow of Engineering New Zealand and an IEEE Senior Member.
\end{IEEEbiographynophoto}

\begin{IEEEbiographynophoto}{Ke Tang (Fellow, IEEE)} received the PhD degree in computer science from Nanyang Technological University, Singapore, in 2007. He is currently a chair professor with the Department of Computer Science and Engineering, Southern University of Science and Technology (SUSTech), Shenzhen, China. His research interests mainly include evolutionary computation, machine learning, and their applications. He was the recipient of the Newton Advanced Fellowship (Royal Society) and the Changjiang Professorship Ministry of Education (MOE) of China, and IEEE Computational Intelligence Society Outstanding Early Career Award and the Natural Science Award of MOE of China. He has been actively involved in the research community, as an organizer/co-organizer of numerous conferences, an associate editor of the IEEE Transactions on Evolutionary Computation, and as a member of Editorial Boards for a few other journals.
\end{IEEEbiographynophoto}

\end{document}